\documentclass{article} 
\usepackage{iclr2017_conference,times}
\usepackage{hyperref}
\usepackage{url}
\usepackage{graphicx}
\usepackage{subcaption}
\usepackage[utf8]{inputenc}
\usepackage{natbib}

\title{generalizable features from Unsupervised learning}

\author{Mehdi Mirza \& Aaron Courville \& Yoshua Bengio\\
MILA \\
Université de Montréal\\
\texttt{\{memirzamo, aaron.courville, yoshua.umontreal\}@gmail.com} \\
}

\begin{document}

\maketitle

\begin{abstract}

Humans learn a predictive model of the world and use this model to reason about future events and the consequences of actions. In contrast to most machine predictors,  we exhibit an impressive ability to generalize to unseen scenarios and reason intelligently in these settings. One important aspect of this ability is \emph{physical intuition} \citep{lake2016building}. In this work, we explore the potential of unsupervised learning to find features that promote better generalization to settings outside the supervised training distribution. Our task is predicting the stability of towers of square blocks. We demonstrate that an unsupervised model, trained to predict future frames of a video sequence of stable and unstable block configurations, can yield features  that support extrapolating stability prediction to blocks configurations outside the training set distribution.
\end{abstract}

\section{Introduction}
Humans learn a tremendous amount of knowledge about the world with almost no supervision and can construct a predictive model of the world. We use this model of the world to interact with our environment. As also argued by \cite{lake2016building} one of the core ingredients of human intelligence is intuitive physics. Children can learn and predict some of the common physical behaviors of our world just by observing and interacting without any direct supervision. And they form a sophisticated predictive model of the physical environment and expect the world to behave based on their mental model and  have a reasonable expectation about unseen situations \cite{teglas2011pure}.

Despite impressive progress in the last few years in the training of the supervised models, we have not yet quite been able to achieve similar results in unsupervised learning, and it remains one of the challenging research areas in the field. The full potential of the application of unsupervised learning is yet to be realized. 

In this work, we leverage unsupervised learning to train a predictive model over sequences. We use the imagined and predicted future sequence data to help a physical environment prediction model generalize better to unseen settings.

More specifically we focus on the task of predicting if a tower of square bricks will fall or not, as introduced by \citet{lerer2016learning}. They showed that a deep convolution neural network could predict the fall of the towers with super-human accuracy. But despite the strengths of convolution neural networks, \citet{zhang2016comparative} shows how deep neural networks have a hard time generalizing to novel situations in the same way as humans or simulation-based models can do. In this work, we show that deep neural networks are capable of generalizing to novel situations through a form of unsupervised learning. The core idea is to observe the world without any supervision and build a future predictive model of it, and in a later stage leverage and utilize the imagined future to train a better fall prediction model.

\section{Related Work}
In the beginning, unsupervised learning and generative models emerged as pre-training method \cite{hinton2006reducing, hinton2006fast, bengio2007greedy} to help other tasks such as supervised learning. But since \cite{krizhevsky2012imagenet} many other regularization \cite{srivastava2014dropout}, weight initialization \cite{glorot2010understanding} and normalization \cite{ioffe2015batch} techniques and architecture designs \cite{he2015deep} has been introduced  that diminish the effect of pre-training. Although pre-training still could be useful in data scarce domains they are many other ways and applications that unsupervised learning are still very interesting models and it is a very active area of research. Just to name a few applications are semi-supervised learning \cite{kingma2014semi, salimans2016improved, dumoulin2016adversarially} super resolution \cite{sonderby2016amortised}.

Video generation is one active area of research with many applications, and many of the recent works have been using some of the states of the art neural networks for video generation. \citet{srivastava2015unsupervised} uses LSTM recurrent neural networks to train an unsupervised future predictive model for video generation. And here we use a very similar architecture as described in Section~\ref{sec:unsupervised}. \citet{mathieu2015deep} combines the common mean-squared-error objective function with an adversarial training cost in order to generate sharper samples. \citet{lotter2016deep} introduce another form of unsupervised video prediction training scheme that manages to predict future events such as the direction of the turn of a car which could have potential use in training of the self-driving cars. 

Model-based reinforcement learning (RL) is an active research area that holds the promise of making the RL agents less data hungry. Learning agents could explore, learn in an unsupervised way about their world, and learn even more by dreaming about future states. We believe that action-condition video prediction models are an important ingredient for this task. \citet{fragkiadaki2015learning} learn the dynamics of billiards balls by supervised training of a neural net. Action-conditioned video prediction models have been applied to Atari playing agent \citet{oh2015action} as well as robotics \citep{finn2016unsupervised, finn2016deep}.


\section{Dataset}
Recent datasets for predicting the stability of block configurations \citep{lerer2016learning, zhang2016comparative} only provide binary labels of stability, and exclude the video simulation of the block configuration. We, therefore, construct a new dataset, with a similar setup as \cite{lerer2016learning, zhang2016comparative}, that includes this video sequence. We use a Javascript based physics engine\footnote{https://chandlerprall.github.io/Physijs/} to generate the data.  

We construct towers made of $3-5$ square blocks. To sample a random tower configuration, we uniformly shift each block in its $x$ and $y$ position such that it touches the block below. Because taller towers are more unstable, this shift is smaller when we add more blocks. To simplify our learning setting, we balance the number of stable and unstable block configurations. For each tower height, we create $8000$, $1000$ and $3000$ video clips for the training, validation, and test set, respectively. The video clips are sub-sampled in time to include more noticeable changes in the blocks configurations. We decided to keep 39 number of frames which with our sub-sampling rate was enough time for unstable towers to collapse. Each video frame is an RGB image of size 64x64. In addition to binary stability label, we include the number of blocks that fell down. 

\section{Architecture}
The core idea of this paper is to use future state predictions of a generative video model to enhance the performance of a supervised prediction model. Our architecture consists of two separate modules:
\begin{description}
\item[Frame predictor] A generative model to predict future frames of a video sequence. This model is trained to either generate the last frame or the complete sequence of frames. 
\item[Stability predictor] 
In the original task, stability is predicted from a static image of a block configuration. We explore whether, in addition to initial configuration, the last frame prediction of our unsupervised model improves the performance of the stability prediction. 
\end{description}
In the following sections, we explore several different architectures for both modules. 

\subsection{Future frame prediction} \label{sec:unsupervised}
\begin{figure}[ht]
\begin{center}
 \includegraphics[width=\textwidth]{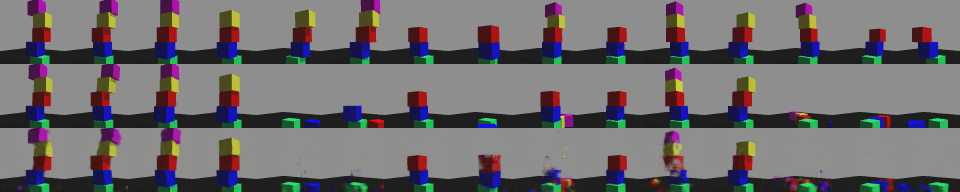}
\end{center}
\caption{Samples from the ConvDeconv model. First and second rows show first and last frame respectively from the test data. And the third row shows generated last frame samples.}
 \label{fig:convdeconv_samples}
\end{figure}

We consider two different model architectures for this task. The first one, named ConvDeconv, 
only takes the first frame as input and predicts the last frame of the video sequence. 
The architecture consist of a block of convolution and max-pooling layers.
To compensate for the dimensionality reduction of the max-pooling layers, we have a fully-connected layer following the last max-pooling layer.
And finally a subsequent block of deconvolution layers with the output size same as the model input size. 
All activation functions are ReLU\citep{nair2010rectified}. See Table~\ref{table:conv_deconv} for more details of the architecture.
The objective function is the mean squared error between the generated last frame and the ground-truth frame; 
as a result, this training will not require any labels. We also experimented with an additional adversarial 
cost as in \cite{mathieu2015deep} but did not observe any improvement for the stability prediction task. 
We hypothesize that although the adversarial objective function helps to have sharper images, such improved 
sample quality does not transfer to better stability prediction. 
Figure~\ref{fig:convdeconv_samples} shows a few examples of the generated data on the test set. 
Mean squared error is minimized using the AdaM Optimizer\citep{kingma2014adam} 
and we use early-stopping when the validation loss does not improve for $100$ epochs.

\begin{figure}[ht]
\begin{center}
 \includegraphics[ height=0.25\textheight]{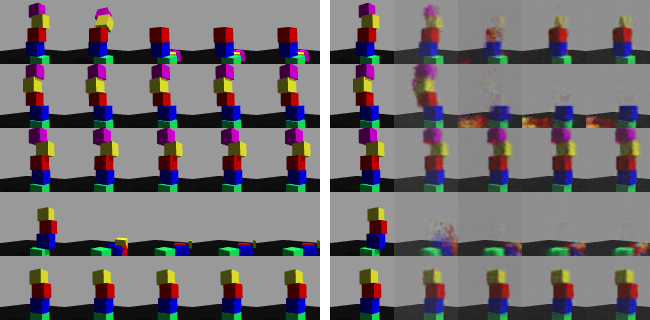}
\end{center}
\caption{Samples from the ConvLSTMDeconv model. Each row is a different sample. The left sequence is the data and the right sequence is the generated data. Note that during generation model only see the first frame and for next time steps uses its own output from the last timestep.}
 \label{fig:lstm_samples}
\end{figure}

\begin{figure}[ht]
\begin{center}
 \includegraphics[width=\textwidth]{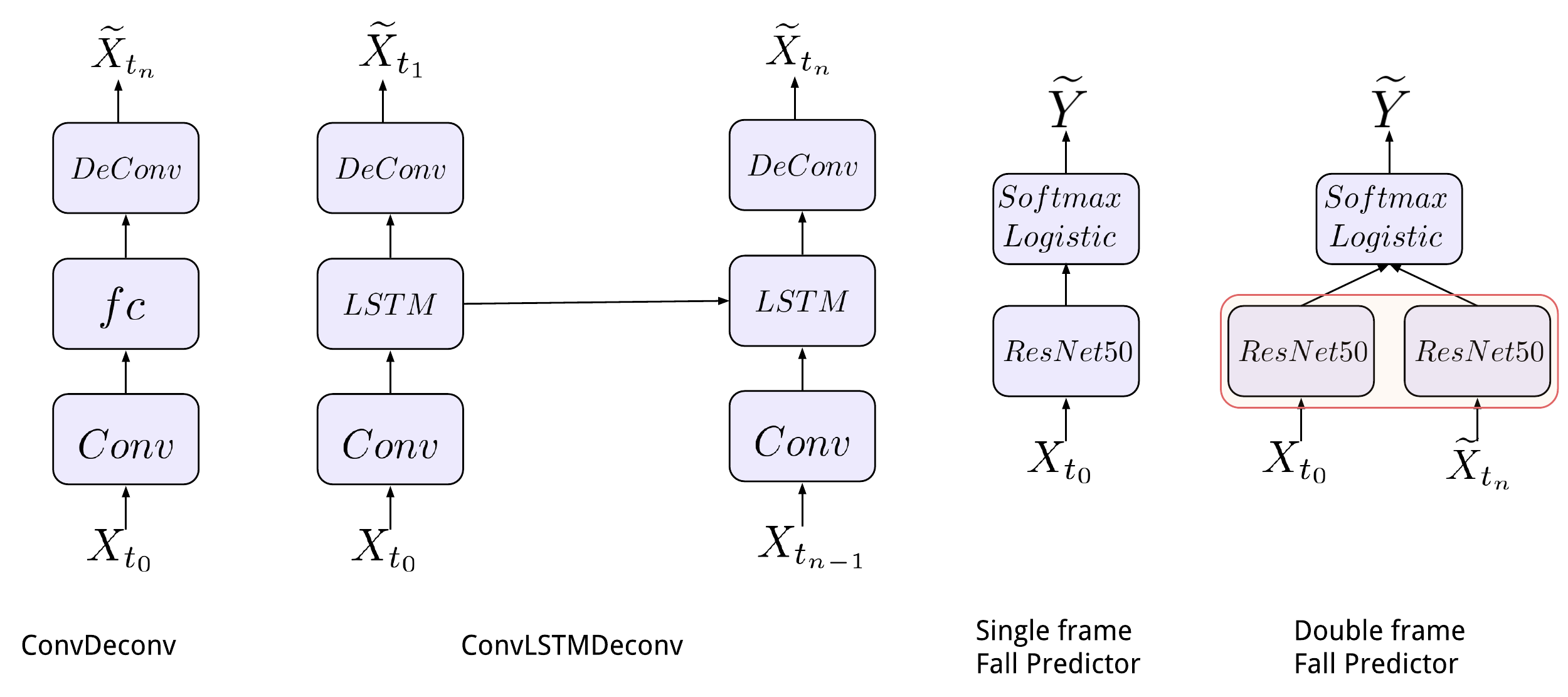}
\end{center}
\caption{Different model architectures. The first two on the left are ConvDeconv and ConvLSTMDeconv described in Section~\ref{sec:unsupervised}. And the two on the right are models used for the supervised fall prediction described in Section~\ref{sec:supervised}. Single frame predictor is the baseline model. And the double frame predictor is the model that uses the generated data.}
 \label{fig:models}
\end{figure}

We extend this ConvDeconv model in a second architecture, named ConvLSTMDeconv, to predict the next frame at each timestep. This model is composed of an LSTM architecture. The same convolutional and deconvolutional blocks as ConvDeconv is utilized to respectively input the current frame to the LSTM transition and output the next frame from the current LSTM state. The details of the ConvLSTMDeconv model architecture are shown in Table~\ref{table:conv_lstm_deconv} and Figure~\ref{fig:models} shows the diagram of the both architectures. During the training at each time step the ground-truth data feeds in to the model, but during the test time only the initial time step gets the first frame from the data and for subsequent time steps the generated frames from the previous time steps feed in to the model. The is similar setup to recurrent neural network language models \cite{mikolov2012statistical}, and this is necessary as during the test time we only have access to the first frame.
As before, the model is trained to predict the next frame at each time step by minimizing the predictive mean-squared-error using AdaM optimizer and early-stopping. For training, we further subsample in time dimension and reduce the sequence length to 5-time steps. Figure~\ref{fig:lstm_samples} shows some sample generated sequences from the test set. 



\begin{table}[!htb]
    \begin{minipage}{.5\linewidth}
      \centering
      \scalebox{0.66}{
        \begin{tabular}{|c | c | c| c| } 
 \hline
 Layer & Type & Output channels/dimensions & Kernel/Pool size \\ [0.5ex] 
 \hline\hline
 1 & Conv & $64$ & $3 \times 3$ \\
  \hline
 2 & MaxPool & $64$ & $4 \times 4$ \\
  \hline
 3 & Conv & $128$ & $3 \times 3$ \\
  \hline
 4 & MaxPool & $64$ & $3 \times 3$ \\
  \hline 
 5 & Conv & $64$ & $3 \times 3$ \\
  \hline
 6 & MaxPool & $64$ & $3 \times 3$ \\
  \hline
 7 & FC & $64 \times 64 \times 16 = 65536$ &  \\
  \hline
 8 & DeConv & $64$ & $3 \times 3$ \\
  \hline
 9 & DeConv & $128$ & $3 \times 3$ \\
  \hline
 10 & DeConv & $64$ & $3 \times 3$ \\
  \hline
 11 & DeConv & $3$ & $3 \times 3$ \\
 \hline
\end{tabular}}
      \caption{ConvDeconv model architecture. \\FC stands for "Fully Connected".}
      \label{table:conv_deconv}
    \end{minipage}%
    \begin{minipage}{.5\linewidth}
      \centering
      \scalebox{0.66}{
        \begin{tabular}{|c | c | c| c| } 
 \hline
 Layer & Type & Output channels/Dimension & Kernel/Pool size \\ [0.5ex] 
 \hline\hline
 1 & Conv & $64$ & $3 \times 3$ \\
  \hline
 2 & MaxPool & $64$ & $4 \times 4$ \\
  \hline
 3 & Conv & $128$ & $3 \times 3$ \\
  \hline
 4 & MaxPool & $64$ & $3 \times 3$ \\
  \hline 
 5 & Conv & $64$ & $3 \times 3$ \\
  \hline
 6 & MaxPool & $64$ & $3 \times 3$ \\
  \hline
 7 & FC LSTM & $2000$ &  \\
  \hline
 8 & FC & $64 \times 64 \times 3$ &  \\
  \hline
 9 & DeConv & $64$ & $3 \times 3$ \\
  \hline
 10 & DeConv & $64$ & $3 \times 3$ \\
  \hline
 11 & DeConv & $3$ & $3 \times 3$ \\
 \hline
\end{tabular}}
      \caption{ConvLSTMDeconv model architecture. FC stands for "Fully Connected".}
      \label{table:conv_lstm_deconv}
    \end{minipage} 
\end{table}

\subsection{Stability prediction} \label{sec:supervised}
We have two supervised models for stability prediction. The first one will be a baseline that takes as input the first frame and predict the fall of the tower. For this model we use 50 layer ResNet architecture from \cite{he2016identity}. We trained the baseline model on each of the different tower heights {3, 4, 5}. We call it the single model and name experiments {3S, 4S, 5S} respectively for the number of blocks that it was trained on.
The second model will be the one using the generated data: it takes as input the first frame and the generated last frame. It consisted of two 50 Layer ResNet blocks in parallel, one for the first frame and one for last frame and the last hidden layer of both models are concatenated together before a logistic regression layer (or Softmax in the case of non-binary labels). Both ResNet blocks share parameters. Based on whether the generated data is coming from ConvDeconv model or ConvLSTMDeconv model we labeled experiments as {3CD, 4CD, 5CD} and {3CLD, 4CLD, 5CLD} respectively.

None of the models are pre-trained and all the weights are randomly initialized. As in \ref{sec:unsupervised}, we use AdaM and we stopped the training when the validation accuracy was not improved for $100$ epochs. All images are contrast normalized independently and we augment our training set using random horizontal flip of the images and randomly changing the contrast and brightness.

\section{Results}
\begin{figure}[ht]
\begin{center}
 \includegraphics[width=\textwidth]{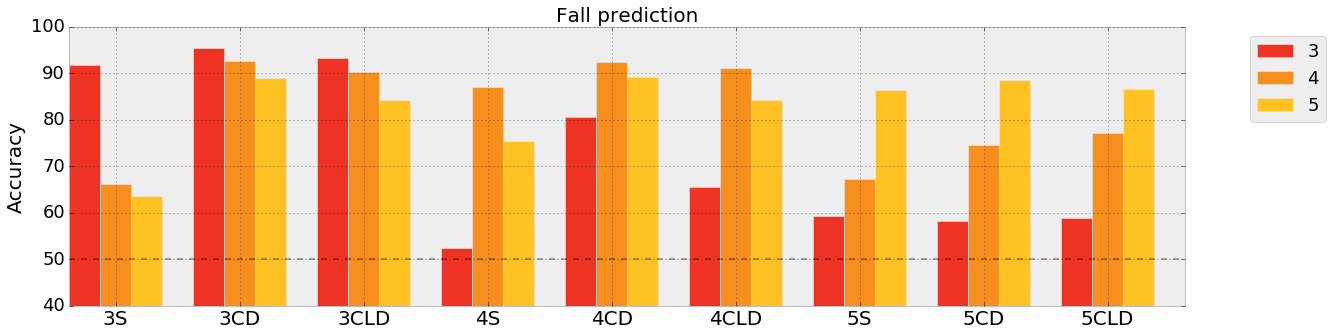}
\end{center}
\caption{Accuracy in percentage for each of the 9 models tested on test sets with a different number of blocks. Each color represents the number of blocks that the model was tested on. $50\%$ is chance.}
 \label{fig:binary_label}
\end{figure}

Figure~\ref{fig:binary_label} shows the classification results for each of the 9 models described in Section~\ref{sec:supervised} tested on 3, 4 and 5 blocks. Each test case is shown with a different color. And Table~\ref{table:results} shows all the 27 test case results' numerical values. In almost all cases the generated data improves the generalization performance to test cases with a different number of blocks than it was trained on. For comparison we have included results from \cite{zhang2016comparative} in Table~\ref{table:zhang_results}. Since \cite{zhang2016comparative} only reports the results when the models are trained on tower of 4 blocks, the corresponding results would be the second block row in Table~\ref{table:results}, models 4S, 4CD and 4CLD. Even though the datasets are not the same, but it can be observed that the range of performance of the baseline 4S model is consistent with the range of performance of AlexNet model on Table~\ref{table:zhang_results}. It can be seen that how the results of the 4CD model are significantly better than both IPE and human performance reported in \cite{zhang2016comparative}, while the baselines have similar performances. 

One observation is the fact that the improvements are more significant when it's been tested on scenarios with more bricks than during training. It also improves the reverse case, i.e. fewer bricks than during training, but the improvement is not as significant. It is worth mentioning that testing on a lower number of bricks is a much harder problem as pointed out in \cite{zhang2016comparative} too. In their case, the prediction performance was almost random when going from 4 blocks to 3 blocks, which is not the case in our experiments\footnote{We are not using the same dataset as \cite{zhang2016comparative} and hence direct comparison is not possible.}. One possible explanation for performance loss is that a balanced tower with fewer blocks corresponds to an unstable configuration for a tower with more blocks e.g. a tower with 3 blocks is classified as unstable for a prediction model trained on towers of 5 blocks. One solution could be to train these models to predict how many blocks have fallen instead of a binary stability label. Because we have access to this data in our dataset, we explored the same experiments using these labels. Unfortunately, we did not observe any significant improvement. The main reason could be that the distribution of the number of fallen blocks is extremely unbalanced. It is hard to collect data with a balanced number of fallen blocks because some configurations are thus very unlikely e.g. a tower of 5 blocks with only two blocks falls (the majority of the time the whole tower collapses).

The another observation is the fact that models that use ConvDeconv generated data performed slightly better than those that use ConvLSTMDeconv. As seen in Figure~\ref{fig:lstm_samples} the samples in the ConvLSTMDeconv case are more noisy and less sharper than those in Figure~\ref{fig:convdeconv_samples}. This could be caused since after the first time step the model outputs from the last time step is used as input for the next time step, the samples degenerates the longer the sequence is. 

Data augmentation was crucial to increase the generalization performance of the stability prediction e.g. 5CD model tested on 4 bricks achieved only $50\%$ without data augmentation while reaching $74.5 \%$ accuracy with data augmentation. This significant improvement from data augmentation could be partly because our dataset was relatively small.

\begin{table}[!htb]
    \begin{minipage}{.5\linewidth}
      \centering
      \scalebox{0.8}{
        \begin{tabular}{| c | c | c | c | }
\hline
 Model & Train set & Test set & Accuracy \\
 \hline
 3S & 3 & 3 & 91.87 \%\\
 3S & 3 & 4 & 66.1 \%\\
 3S & 3 & 5 & 63.7 \%\\
 
 3CD & 3 & 3 & 95.5 \%\\
 3CD & 3 & 4 & 92.63 \%\\
 3CD & 3 & 5 & 89 \%\\
 
 3CLD & 3 & 3 & 93.3 \%\\
 3CLD & 3 & 4 & 90.33 \%\\
 3CLD & 3 & 5 & 84.30 \%\\
 \hline
 
 4S & 4 & 3 & 52.5 \%\\
 4S & 4 & 4 & 87 \%\\
 4S & 4 & 5 & 75.53 \%\\
 
 4CD & 4 & 3 & 80.53 \%\\
 4CD & 4 & 4 & 92.5 \%\\
 4CD & 4 & 5 & 89.1 \%\\
 
 4CLD & 4 & 3 & 65.53 \%\\
 4CLD & 4 & 4 & 91.20 \%\\
 4CLD & 4 & 5 & 84.20 \%\\
\hline 

 5S & 5 & 3 & 59.26 \%\\
 5S & 5 & 4 & 67.23 \%\\
 5S & 5 & 5 & 86.50 \%\\
 
 5CD & 5 & 3 & 58.27 \%\\
 5CD & 5 & 4 & 74.50 \%\\
 5CD & 5 & 5 & 88.53 \%\\
 
 5CLD & 5 & 3 & 58.90 \%\\
 5CLD & 5 & 4 & 74.50 \%\\
 5CLD & 5 & 5 & 88.53 \%\\
 \hline 
\end{tabular}}
\caption{The results from our experiments}
\label{table:results}
    \end{minipage}%
    \begin{minipage}{.5\linewidth}
      \centering
        \scalebox{0.8}{
        \begin{tabular}{| c | c | c | c | }
\hline
 Model & Train set & Test set & Accuracy \\
 \hline
 
 AlexNet & 4 & 3 & 51 \%\\
 AlexNet & 4 & 4 & 95 \%\\
 AlexNet & 4 & 5 & 78.5 \%\\
 \hline 
 IPE & N/A & 3 & 72 \%\\
 IPE & N/A & 4 & 64 \%\\
 IPE & N/A & 5 & 56 \%\\
 \hline
 Human & N/A & 3 & 76.5 \%\\
 Human & N/A & 4 & 68.5 \%\\
 Human & N/A & 5 & 59 \%\\
 \hline 
\end{tabular}}
    \caption{The results reported on \cite{zhang2016comparative}. We emphasize that these results are on a different dataset.}
    \label{table:zhang_results}
    \end{minipage} 
\end{table}

\section{Conclusion}
In this paper, we showed that data generated from an unsupervised model could help a supervised learner to generalize to unseen scenarios.
We argue that this ability of transfer learning and generalization by observing the world could be one of the ingredients to construct
a model of the world that could have applications in many tasks, such as model-based RL. We aim to extend this work in future by looking
at the videos of robots manipulating objects and being able to predict their failure beforehand, which could help an RL agent to explore
more intelligently.

\subsubsection*{Acknowledgments}
We would like to thank Harm de Vries and Laurent Dinh for their help and feedback in writing the paper. And also thank Adam Lerer 
and Jiajun Wu for sharing their dataset. We thank NSERC, CIFAR, IBM, Canada Research Chairs, Google and Samsung for funding.


\bibliography{iclr2017_conference}

\begin{thebibliography}{27}
\providecommand{\natexlab}[1]{#1}
\providecommand{\url}[1]{\texttt{#1}}
\expandafter\ifx\csname urlstyle\endcsname\relax
  \providecommand{\doi}[1]{doi: #1}\else
  \providecommand{\doi}{doi: \begingroup \urlstyle{rm}\Url}\fi

\bibitem[Bengio et~al.(2007)Bengio, Lamblin, Popovici, Larochelle,
  et~al.]{bengio2007greedy}
Yoshua Bengio, Pascal Lamblin, Dan Popovici, Hugo Larochelle, et~al.
\newblock Greedy layer-wise training of deep networks.
\newblock \emph{Advances in neural information processing systems},
  19:\penalty0 153, 2007.

\bibitem[Dumoulin et~al.(2016)Dumoulin, Belghazi, Poole, Lamb, Arjovsky,
  Mastropietro, and Courville]{dumoulin2016adversarially}
Vincent Dumoulin, Ishmael Belghazi, Ben Poole, Alex Lamb, Martin Arjovsky,
  Olivier Mastropietro, and Aaron Courville.
\newblock Adversarially learned inference.
\newblock \emph{arXiv preprint arXiv:1606.00704}, 2016.

\bibitem[Finn \& Levine(2016)Finn and Levine]{finn2016deep}
Chelsea Finn and Sergey Levine.
\newblock Deep visual foresight for planning robot motion.
\newblock \emph{arXiv preprint arXiv:1610.00696}, 2016.

\bibitem[Finn et~al.(2016)Finn, Goodfellow, and Levine]{finn2016unsupervised}
Chelsea Finn, Ian Goodfellow, and Sergey Levine.
\newblock Unsupervised learning for physical interaction through video
  prediction.
\newblock \emph{arXiv preprint arXiv:1605.07157}, 2016.

\bibitem[Fragkiadaki et~al.(2015)Fragkiadaki, Agrawal, Levine, and
  Malik]{fragkiadaki2015learning}
Katerina Fragkiadaki, Pulkit Agrawal, Sergey Levine, and Jitendra Malik.
\newblock Learning visual predictive models of physics for playing billiards.
\newblock \emph{arXiv preprint arXiv:1511.07404}, 2015.

\bibitem[Glorot \& Bengio(2010)Glorot and Bengio]{glorot2010understanding}
Xavier Glorot and Yoshua Bengio.
\newblock Understanding the difficulty of training deep feedforward neural
  networks.
\newblock In \emph{Aistats}, volume~9, pp.\  249--256, 2010.

\bibitem[He et~al.(2015)He, Zhang, Ren, and Sun]{he2015deep}
Kaiming He, Xiangyu Zhang, Shaoqing Ren, and Jian Sun.
\newblock Deep residual learning for image recognition.
\newblock \emph{arXiv preprint arXiv:1512.03385}, 2015.

\bibitem[He et~al.(2016)He, Zhang, Ren, and Sun]{he2016identity}
Kaiming He, Xiangyu Zhang, Shaoqing Ren, and Jian Sun.
\newblock Identity mappings in deep residual networks.
\newblock \emph{arXiv preprint arXiv:1603.05027}, 2016.

\bibitem[Hinton \& Salakhutdinov(2006)Hinton and
  Salakhutdinov]{hinton2006reducing}
Geoffrey~E Hinton and Ruslan~R Salakhutdinov.
\newblock Reducing the dimensionality of data with neural networks.
\newblock \emph{Science}, 313\penalty0 (5786):\penalty0 504--507, 2006.

\bibitem[Hinton et~al.(2006)Hinton, Osindero, and Teh]{hinton2006fast}
Geoffrey~E Hinton, Simon Osindero, and Yee-Whye Teh.
\newblock A fast learning algorithm for deep belief nets.
\newblock \emph{Neural computation}, 18\penalty0 (7):\penalty0 1527--1554,
  2006.

\bibitem[Ioffe \& Szegedy(2015)Ioffe and Szegedy]{ioffe2015batch}
Sergey Ioffe and Christian Szegedy.
\newblock Batch normalization: Accelerating deep network training by reducing
  internal covariate shift.
\newblock \emph{arXiv preprint arXiv:1502.03167}, 2015.

\bibitem[Kingma \& Ba(2014)Kingma and Ba]{kingma2014adam}
Diederik Kingma and Jimmy Ba.
\newblock Adam: A method for stochastic optimization.
\newblock \emph{arXiv preprint arXiv:1412.6980}, 2014.

\bibitem[Kingma et~al.(2014)Kingma, Mohamed, Rezende, and
  Welling]{kingma2014semi}
Diederik~P Kingma, Shakir Mohamed, Danilo~Jimenez Rezende, and Max Welling.
\newblock Semi-supervised learning with deep generative models.
\newblock In \emph{Advances in Neural Information Processing Systems}, pp.\
  3581--3589, 2014.

\bibitem[Krizhevsky et~al.(2012)Krizhevsky, Sutskever, and
  Hinton]{krizhevsky2012imagenet}
Alex Krizhevsky, Ilya Sutskever, and Geoffrey~E Hinton.
\newblock Imagenet classification with deep convolutional neural networks.
\newblock In \emph{Advances in neural information processing systems}, pp.\
  1097--1105, 2012.

\bibitem[Lake et~al.(2016)Lake, Ullman, Tenenbaum, and
  Gershman]{lake2016building}
Brenden~M Lake, Tomer~D Ullman, Joshua~B Tenenbaum, and Samuel~J Gershman.
\newblock Building machines that learn and think like people.
\newblock \emph{arXiv preprint arXiv:1604.00289}, 2016.

\bibitem[Lerer et~al.(2016)Lerer, Gross, and Fergus]{lerer2016learning}
Adam Lerer, Sam Gross, and Rob Fergus.
\newblock Learning physical intuition of block towers by example.
\newblock \emph{arXiv preprint arXiv:1603.01312}, 2016.

\bibitem[Lotter et~al.(2016)Lotter, Kreiman, and Cox]{lotter2016deep}
William Lotter, Gabriel Kreiman, and David Cox.
\newblock Deep predictive coding networks for video prediction and unsupervised
  learning.
\newblock \emph{arXiv preprint arXiv:1605.08104}, 2016.

\bibitem[Mathieu et~al.(2015)Mathieu, Couprie, and LeCun]{mathieu2015deep}
Michael Mathieu, Camille Couprie, and Yann LeCun.
\newblock Deep multi-scale video prediction beyond mean square error.
\newblock \emph{arXiv preprint arXiv:1511.05440}, 2015.

\bibitem[Mikolov(2012)]{mikolov2012statistical}
Tom{\'a}{\v{s}} Mikolov.
\newblock Statistical language models based on neural networks.
\newblock \emph{Presentation at Google, Mountain View, 2nd April}, 2012.

\bibitem[Nair \& Hinton(2010)Nair and Hinton]{nair2010rectified}
Vinod Nair and Geoffrey~E Hinton.
\newblock Rectified linear units improve restricted boltzmann machines.
\newblock In \emph{Proceedings of the 27th International Conference on Machine
  Learning (ICML-10)}, pp.\  807--814, 2010.

\bibitem[Oh et~al.(2015)Oh, Guo, Lee, Lewis, and Singh]{oh2015action}
Junhyuk Oh, Xiaoxiao Guo, Honglak Lee, Richard~L Lewis, and Satinder Singh.
\newblock Action-conditional video prediction using deep networks in atari
  games.
\newblock In \emph{Advances in Neural Information Processing Systems}, pp.\
  2863--2871, 2015.

\bibitem[Salimans et~al.(2016)Salimans, Goodfellow, Zaremba, Cheung, Radford,
  and Chen]{salimans2016improved}
Tim Salimans, Ian Goodfellow, Wojciech Zaremba, Vicki Cheung, Alec Radford, and
  Xi~Chen.
\newblock Improved techniques for training gans.
\newblock \emph{arXiv preprint arXiv:1606.03498}, 2016.

\bibitem[S{\o}nderby et~al.(2016)S{\o}nderby, Caballero, Theis, Shi, and
  Husz{\'a}r]{sonderby2016amortised}
Casper~Kaae S{\o}nderby, Jose Caballero, Lucas Theis, Wenzhe Shi, and Ferenc
  Husz{\'a}r.
\newblock Amortised map inference for image super-resolution.
\newblock \emph{arXiv preprint arXiv:1610.04490}, 2016.

\bibitem[Srivastava et~al.(2014)Srivastava, Hinton, Krizhevsky, Sutskever, and
  Salakhutdinov]{srivastava2014dropout}
Nitish Srivastava, Geoffrey~E Hinton, Alex Krizhevsky, Ilya Sutskever, and
  Ruslan Salakhutdinov.
\newblock Dropout: a simple way to prevent neural networks from overfitting.
\newblock \emph{Journal of Machine Learning Research}, 15\penalty0
  (1):\penalty0 1929--1958, 2014.

\bibitem[Srivastava et~al.(2015)Srivastava, Mansimov, and
  Salakhutdinov]{srivastava2015unsupervised}
Nitish Srivastava, Elman Mansimov, and Ruslan Salakhutdinov.
\newblock Unsupervised learning of video representations using lstms.
\newblock \emph{CoRR, abs/1502.04681}, 2, 2015.

\bibitem[T{\'e}gl{\'a}s et~al.(2011)T{\'e}gl{\'a}s, Vul, Girotto, Gonzalez,
  Tenenbaum, and Bonatti]{teglas2011pure}
Ern{\H{o}} T{\'e}gl{\'a}s, Edward Vul, Vittorio Girotto, Michel Gonzalez,
  Joshua~B Tenenbaum, and Luca~L Bonatti.
\newblock Pure reasoning in 12-month-old infants as probabilistic inference.
\newblock \emph{science}, 332\penalty0 (6033):\penalty0 1054--1059, 2011.

\bibitem[Zhang et~al.(2016)Zhang, Wu, Zhang, Freeman, and
  Tenenbaum]{zhang2016comparative}
Renqiao Zhang, Jiajun Wu, Chengkai Zhang, William~T Freeman, and Joshua~B
  Tenenbaum.
\newblock A comparative evaluation of approximate probabilistic simulation and
  deep neural networks as accounts of human physical scene understanding.
\newblock \emph{arXiv preprint arXiv:1605.01138}, 2016.

\end{thebibliography}
\bibliographystyle{iclr2017_conference}

\end{document}